\def\year{2020}\relax
\documentclass[letterpaper]{article} 
\usepackage{aaai20}  
\usepackage{times}  
\usepackage{helvet} 
\usepackage{courier}  
\usepackage[hyphens]{url}  
\usepackage{graphicx} 
\usepackage{graphicx}  
\usepackage{booktabs}
\usepackage{bm}
\usepackage{amsmath}
\usepackage{amssymb}
\usepackage{algorithmicx,algorithm}
\usepackage[noend]{algpseudocode}
\usepackage{multirow}
\newcommand{\citet}[1]{\citeauthor{#1} \shortcite{#1}} 
\newcommand{\citep}{\cite}

\title{Minimizing the Bag-of-Ngrams Difference for Non-Autoregressive \\Neural Machine Translation}

\author{Chenze Shao\textsuperscript{\rm 1}\textsuperscript{\rm 2}, Jinchao Zhang\textsuperscript{\rm 3}, Yang Feng\textsuperscript{\rm 1}\textsuperscript{\rm 2}\thanks{Corresponding author: Yang Feng}, Fandong Meng\textsuperscript{\rm 3} \and Jie Zhou\textsuperscript{\rm 3}\\ 
\textsuperscript{\rm 1} Key Laboratory of Intelligent Information Processing,\\ Institute of Computing Technology, Chinese Academy of Sciences (ICT/CAS)\\
\textsuperscript{\rm 2} University of Chinese Academy of Sciences\\
\textsuperscript{\rm 3} Pattern Recognition Center, WeChat AI, Tencent Inc, China\\
{ \{shaochenze18z,fengyang\}@ict.ac.cn}\\
{ \{dayerzhang,fandongmeng,withtomzhou\}@tencent.com} 
}

\begin{document}
\maketitle
\newcommand\blfootnote[1]{%
\begingroup 
\renewcommand\thefootnote{}\footnote{#1}%
\addtocounter{footnote}{-1}%
\endgroup
}
\begin{abstract}

Non-Autoregressive Neural Machine Translation (NAT) achieves significant decoding speedup through generating target words independently and simultaneously. However, in the context of non-autoregressive translation, the word-level cross-entropy loss cannot model the target-side sequential dependency properly, leading to its weak correlation with the translation quality. As a result, NAT tends to generate influent translations with over-translation and under-translation errors. In this paper, we propose to train NAT to minimize the Bag-of-Ngrams (BoN) difference between the model output and the reference sentence. The bag-of-ngrams training objective is differentiable and can be efficiently calculated, which encourages NAT to capture the target-side sequential dependency and correlates well with the translation quality. We validate our approach on three translation tasks and show that our approach largely outperforms the NAT baseline by about 5.0 BLEU scores on WMT14 En$\leftrightarrow$De and about 2.5 BLEU scores on WMT16 En$\leftrightarrow$Ro.
\blfootnote{ Joint work with Pattern Recognition Center, WeChat AI, Tencent Inc, China.}
\blfootnote{\noindent Reproducible code: \url{https://github.com/ictnlp/BoN-NAT}}
\end{abstract}

\section{Introduction}

Neural Machine Translation (NMT) has achieved impressive performance over the recent years~\cite{cho2014learning,sutskever2014sequence,bahdanau2014neural,wu2016google,vaswani2017attention}. NMT models are typically built on the encoder-decoder framework where the encoder encodes the source sentence into distributed representations and the decoder generates the target sentence from the representations in an autoregressive manner: to predict the next word, previously predicted words have to be fed as inputs. The word-by-word generation manner of NMT determined by the autoregressive mechanism leads to high translation latency during inference and restricts application scenarios of NMT.

To reduce the translation latency for neural machine translation, non-autoregressive translation models~\cite{gu2017non} have been proposed. A basic NAT model takes the same encoder-decoder architecture as Transformer~\cite{vaswani2017attention}. Instead of using previously generated tokens as in autoregressive models, NAT takes other global signals as decoder inputs~\cite{gu2017non,guo2019non,kaiser2018fast,akoury2019syntactically}, which enables the simultaneous and independent generation of target words and achieves significant decoding speedup. 
\begin{figure}[t]
  \begin{center}
    \includegraphics[width=.95\columnwidth]{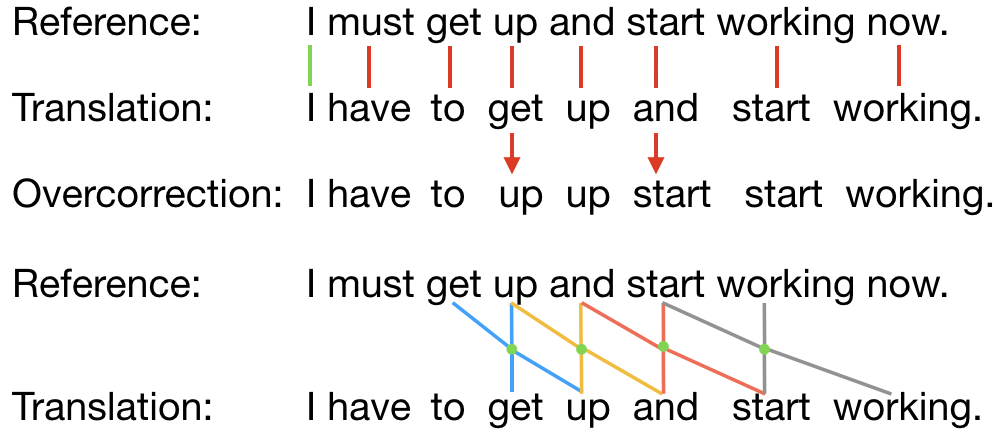}
    \caption{The cross-entropy loss gives a large penalty to the unaligned translation and may cause the overcorrection. The bag-of-2grams difference is robust to the unaligned translation and correlates well with the translation quality.}
    \label{fig:overcorrection}
  \end{center}
\end{figure}

However, in the context of non-autoregressive translation, the word-level cross-entropy loss cannot model the target-side sequential dependency properly, leading to its weak correlation with the translation quality. The cross-entropy loss encourages NAT to generate the golden token in each position without considering the global correctness. For example, as Figure \ref{fig:overcorrection} illustrates, though the translation ``I have to get up and start working'' is semantically close to the reference, the cross-entropy loss will give it a large penalty due to its unalignment with the reference. Under the guidance of cross-entropy loss, the translation may be corrected to ``I have to up up start start working'' that is preferred by the cross-entropy loss but actually gets worse and contains repeated words, which is named as the overcorrection error~\cite{zhang-etal-2019-bridging}. The limitation of the cross-entropy loss aggravates the weakness of NAT in modeling target sequential dependency, which often causes influent translation results with over-translation and under-translation errors~\cite{shao2019retrieving,wang2019non}. In addition, since the cross-entropy loss does not correlate well with the translation quality, it usually takes a long time for NAT to converge.

In this paper, we propose to model the target-side sequential dependency for NAT through minimizing the Bag-of-Ngrams (BoN) difference between the model output and the reference sentence. As the word-level cross-entropy loss cannot model the target-side sequential dependency properly, we propose to evaluate the NAT output on the n-gram level. Since the output of NAT may not be aligned with the reference, we do not require the strict alignment and instead optimize the bag-of-ngrams for NAT. As Figure \ref{fig:overcorrection} illustrates, the bag-of-ngrams difference is robust to unaligned translations and correlates well with the translation quality. Optimizing such a sequence-level objective usually faces the difficulty of exponential search space. Previous works~\cite{ranzato2015sequence,shao2019retrieving} usually observe the objective as a reward and train the model under the reinforcement learning framework~\cite{williams1992simple}, which has relatively slow training speed and suffers from high estimation variance. We introduce a novel method to overcome the difficulty of exponential search space. Firstly, we define the BoN of NAT by the expectation of BoN on all possible translations. Then the BoN training objective is to minimize the BoN difference between NAT output and reference. For NAT, we give a simple and efficient method to calculate the BoN objective. The BoN training objective has the following advantages:
\begin{itemize}
\item It models the target-side sequential dependency for NAT and correlates well with the translation quality.
\item It does not assume specific NAT model architectures and is easy to implement.
\item It can be calculated accurately without making any approximation.
\item It is differentiable and maintains fast training speed. Therefore it can not only be utilized for fine-tuning but also work together with cross-entropy loss to jointly train NAT from scratch.
\end{itemize}

We evaluate the proposed method on three translation tasks (IWSLT16 En$\rightarrow$De, WMT14 En$\leftrightarrow$De, WMT16 En$\leftrightarrow$Ro). Experimental results show that the fine-tuning method achieves large improvements over the pre-trained NAT baseline, and the joint training method further brings considerable improvements over the fine-tuning method, which outperforms the NAT baseline by about 5.0 BLEU scores on WMT14 En$\leftrightarrow$De and about 2.5 BLEU scores on WMT16 En$\leftrightarrow$Ro. 

\section{Background}
\subsection{Autoregressive Neural Machine Translation}
Deep neural networks with autoregressive encoder-decoder framework have achieved great success on machine translation, with different choices of architectures such as RNN, CNN and Transformer. RNN-based models~\cite{bahdanau2014neural,cho2014learning} have a sequential architecture that prevents them from being parallelized. CNN~\cite{gehring2017convolutional}, and self-attention~\cite{vaswani2017attention} based models have highly parallelized architecture, which solves the parallelization problem during training. However, during inference, the translation has to be generated word-by-word due to the autoregressive mechanism. 

Given a source sentence $\bm{X}=\{x_1, ..., x_{n}\}$ and a target sentence $\bm{Y}=\{y_1, ..., y_{T}\}$, autoregressive NMT models the transaltion probability from $\bm{X}$ to $\bm{Y}$ sequentially as:
\begin{equation}
\label{eq:auto_prob}
P(\bm{Y}|\bm{X},\theta) = \prod_{t=1}^{T}p(y_t|\bm{y_{<t}},\bm{X},\theta),
\end{equation}
where $\theta$ is a set of model parameters and $\bm{y_{<t}}=\{y_1,\cdots,y_{t-1}\}$ is the translation history. The standard training objective is the cross-entropy loss, which minimize the negative log-likelihood as:

\begin{equation}
\begin{aligned}
\label{eq:auto_mle}
\mathcal{L}_{MLE}(\theta) = -\sum_{t=1}^{T}\log(p( y _t|\bm{ y_}{<t},\bm{X},\theta)),
\end{aligned}
\end{equation}
During decoding, the partial translation generated by decoding algorithms such as the greedy search and beam search is fed into the decoder to generate the next word.

\subsection{Non-Autoregressive Neural Machine Translation}
Non-autoregressive neural machine translation~\cite{gu2017non} is proposed to reduce the translation latency through parallelizing the decoding process. A basic NAT model takes the same encoder-decoder architecture as Transformer. The NAT encoder stays unchanged from the original Transformer encoder, and the NAT decoder has an additional positional attention block in each layer compared to the Transformer decoder. Besides, there is a length predictor that takes encoder states as input to predict the target length. 

Given a source sentence $\bm{X}=\{x_1, ..., x_{n}\}$ and a target sentence $\bm{Y}=\{y_1, ..., y{_T}\}$, NAT models the translation probability from $\bm{X}$ to $\bm{Y}$ as:

\begin{equation}
\label{eq:nonauto_prob}
P(\bm{Y}|\bm{X},\theta) = \prod_{t=1}^{T}p(y_t|\bm{X},\theta),
\end{equation}
where $\theta$ is a set of model parameters and $p(y_t|\bm{X},\theta)$ indicates the translation probability of word $y_t$ in position $t$. The cross-entropy loss is applied to minimize the negative log-likelihood as:
\begin{equation}
\begin{aligned}
\label{eq:nonauto_mle}
\mathcal{L}_{MLE}(\theta) = -\sum_{t=1}^{T}\log(p( y_t|\bm{X},\theta)),
\end{aligned}
\end{equation}
During training, the target length $T$ is usually set as the reference length. During inference, the target length $T$ is obtained from the length predictor, and then the translation of length $T$ is obtained by taking the word with the maximum likelihood at each time step:
\begin{equation}
\label{eq:argmax}
\hat{y_t} = \arg\max_{y_t}p(y_t|\bm{X},\theta).
\end{equation}

\section{Model}
In this section, we will discuss the BoN objective for NAT in detail. We first formulize the BoN of a discrete sentence by the sum of n-gram vectors with one-hot representation. Then we define the BoN of NMT by the expectation of BoN on all possible translations and give an efficient method to calculate the BoN of NAT. The BoN objective is to minimize the difference between the BoN of NAT output and reference, which encourages NAT to capture the target-side sequential dependency and generate high-quality translations.

\subsection{Bag-of-Ngrams}
Bag-of-words~\cite{joachims1998text} is the most commonly used text representation model which discards the word order and represent sentence as the multiset of its belonging words. 
Bag-of-ngrams~\cite{pang2002thumbs,li2017neural} is proposed to enhance the text representation by takeing consecutive words (n-gram) into consideration. Assume that the vocabulary size is $V$, Bag-of-Ngrams (BoN) is a vector of size $V^n$, which is the sum of zero-one vectors where each vector is the one-hot representation of an n-gram. Formally, for a sentence $\bm{Y}=\{y_1, ..., y{_T}\}$, we use $\text{BoN}_{\bm{Y}}$ to denote the bag-of-ngrams of $\bm{Y}$. For an n-gram $\bm{g} = (g_1,\dots,g_n)$, we use $\text{BoN}_{\bm{Y}}(\bm{g})$ to denote the value of $\bm{g}$ in $\text{BoN}_{\bm{Y}}$, which is the number of occurrences of n-gram $\bm{g}$ in sentence $\bm{Y}$ and can be formulized as follows:
\begin{equation}
\label{eq:bon}
\text{BoN}_{\bm{Y}}(\bm{g}) =\sum_{t=0}^{T-n}1\{y_{t+1:t+n} = \bm{g}\},
\end{equation}
where $1\{\cdot\}$ is the indicator function that takes value from $\{0,1\}$ whose value is 1 iff the inside condition holds. 
\begin{figure}[t]
  \begin{center}
    \includegraphics[width=.95\columnwidth]{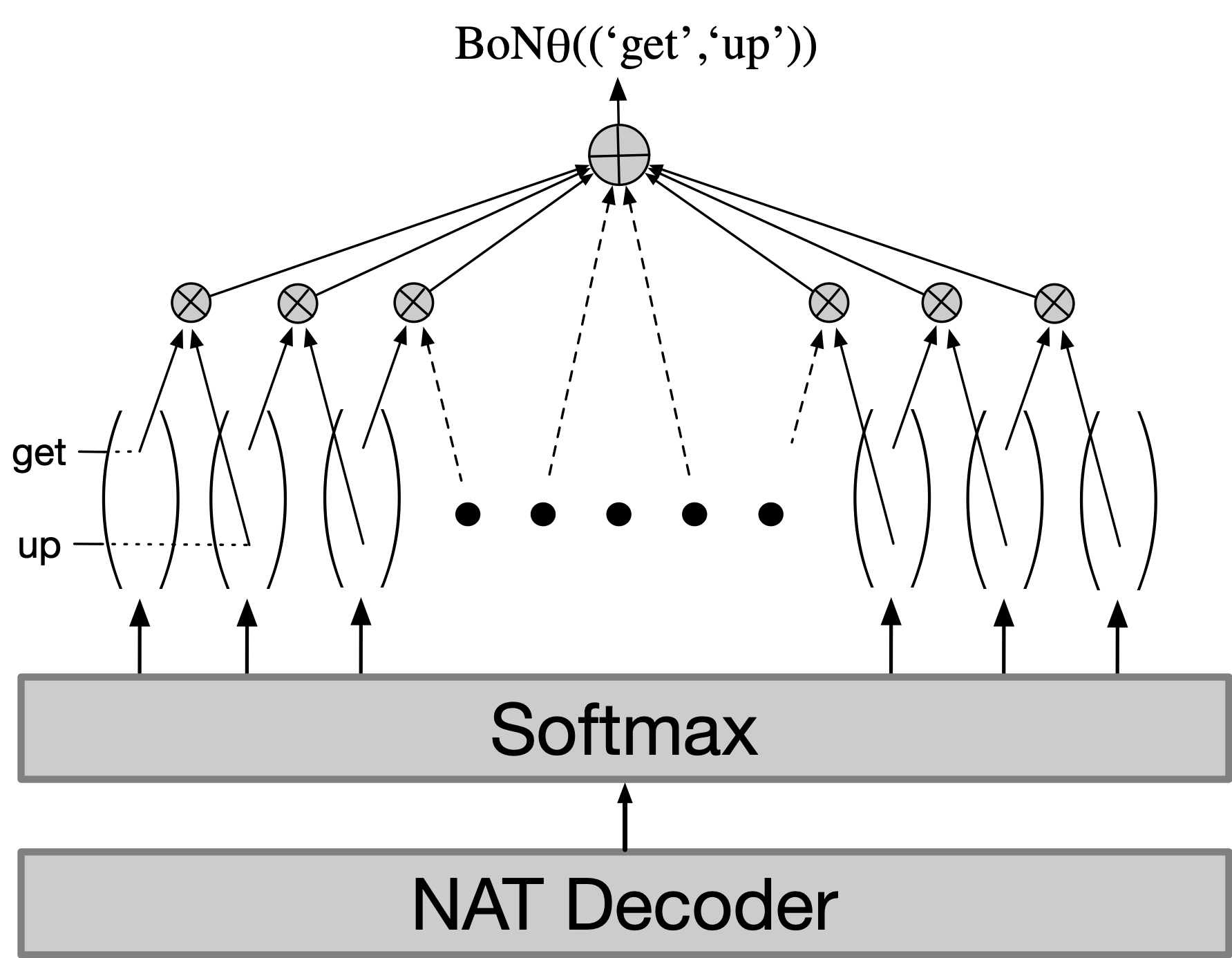}
    \caption{The calculation process of $\text{BoN}_{\theta}$(``get up"). First calaulate the probability of the bigram ``get up" in each subarea and then accumulate the probabilities.}
    \label{fig:bon}
  \end{center}
\end{figure}

For a discrete sentence, our definition of BoN is consistent with previous work. However, there is no clear definition of BoN for sequence models like NMT, which model the probability distribution on the whole target space. A natural approach is to consider all possible translations and use the expected BoN to define the BoN for sequence models. For NMT with parameter $\theta$, we use $\text{BoN}_{\theta}$ to denote its bag-of-ngrams. Formally, given a source sentence $\bm{X}$, the value of n-gram $\bm{g}$ in $\text{BoN}_{\theta}$ is defined as follows:

\begin{equation}
\label{eq:bon_theta}
\text{BoN}_{\theta}(\bm{g}) =\sum_{\bm{Y}} P(\bm{Y}|\bm{X},\theta)\cdot \text{BoN}_{\bm{Y}}(\bm{g}).
\end{equation}

\subsection{Efficient Calculation}
It is unrealistic to directly calculate $\text{BoN}_{\bm{Y}}(\bm{g})$ according to Eq.(\ref{eq:bon_theta}) due to the exponential search space. For autoregressive NMT, because of the conditional dependency in modeling translation probability, it is difficult to simplify the calculation without loss of accuracy. ~\citet{shao2018greedy} only focuses on greedily chosen words and makes the search space very limited. ~\citet{ma2018bag} sums up the distribution of all positions to generate the bag-of-words, which is biased by the teacher forcing. 

Fortunately, NAT models the translation probability in different positions independently, which enables us to divide the target sequence into subareas and analyze the BoN in each subarea without being influenced by other positions. Guiding by this unique property of NAT, we convert Eq.(\ref{eq:bon_theta}) to the following form:

\begin{equation}
\begin{aligned}
\label{eq:bon_theta_reduce}
\text{BoN}_{\theta}&(\bm{g})=\sum_{\bm{Y}}P(\bm{Y}|\bm{X},\theta)\cdot\sum_{t=0}^{T-n}1\{y_{t+1:t+n} = \bm{g}\}\\
=&\sum_{t=0}^{T-n}\sum_{\bm{Y}} P(\bm{Y}|\bm{X},\theta) \cdot 1\{y_{t+1:t+n} = \bm{g}\}\\
=&\sum_{t=0}^{T-n}\sum_{\bm{Y}_{t+1:t+n}} P(\bm{Y}_{t+1:t+n}|\bm{X},\theta) \cdot 1\{y_{t+1:t+n} = \bm{g}\}\\
=&\sum_{t=0}^{T-n}\prod_{i=1}^{n}p(y_{t+i}=g_i|\bm{X},\theta).
\end{aligned}
\end{equation}
Eq.(\ref{eq:bon_theta_reduce}) gives an efficient method to calculate $\text{BoN}_{\theta}(\bm{g})$: slide a window on NAT output distributions to obtain subareas of size $n$, and then accumulate the values of n-gram $\bm{g}$ in all subareas. Figure \ref{fig:bon} illustrates the calculation process of a bigram ``get up". It does not make any approximation and requires little computational effort.

\subsection{Bag-of-Ngrams Objective}
The training objective is to minimize the BoN difference between NAT output and reference. The difference can be measured by several metrics such as the $L_1$ distance, $L_2$ distance and cosine distance. In previous sections, we define the BoN for NAT and give an efficient calculation method for $\text{BoN}_{\theta}(\bm{g})$. However, since there are $V^n$ different n-grams, calculating the complete BoN vector for NAT still consumes a huge amount of storage space and computing resources. 

We use the sparsity of bag-of-ngrams to further simplify the calculation. As shown in Eq.(\ref{eq:bon_theta_reduce}), for NAT, its bag-of-ngrams $\text{BoN}_{\theta}$ is dense. On the contrary, assume that the reference sentence is $\hat{\bm{Y}}$, the vector $\text{BoN}_{\hat{\bm{Y}}}$ is very sparse where only a few entries of it have non-zero values. Using this property, we show that the $L_1$ distance between the two BoN vectors can be calculated efficiently. Unfortunately, some other distance metrics like $L_2$ distance and cosine distance cannot be simplified in this way.

Assume that the target length is $T$. Intutively, a sentence of length $T$ has $T-n+1$ n-grams. We first verify the intution that the $L_1$-norm of $\text{BoN}_{\bm{Y}}$ and $\text{BoN}_{\theta}$ are both $T-n+1$:

\begin{equation}
\label{eq:l1-norm}
\sum_{\bm{g}}\text{BoN}_{\bm{Y}}(\bm{g})=\sum_{t=0}^{T-n}\sum_{\bm{g}}1\{y_{t+1:t+n} = \bm{g}\}=T-n+1.
\end{equation}
\begin{equation}
\begin{aligned}
&\sum_{\bm{g}}\text{BoN}_{\theta}(\bm{g})=\sum_{\bm{g}}\sum_{\bm{Y}} P(\bm{Y}|\bm{X},\theta)\cdot \text{BoN}_{\bm{Y}}(\bm{g})\\
&=\sum_{\bm{Y}} P(\bm{Y}|\bm{X},\theta)\cdot \sum_{\bm{g}}\text{BoN}_{\bm{Y}}(\bm{g})=T-n+1.
\end{aligned}
\end{equation}

On this basis, we can derive the $L_1$ distance between $\text{BoN}_{\theta}$ and $\text{BoN}_{\hat{\bm{Y}}}$, where $\hat{\bm{Y}}$ is the reference sentence. We denote the $L_1$ distance as BoN-$L_1$ and convert it to the following form:
\begin{equation}
\begin{aligned}
&\text{BoN-}L_1=\sum_{\bm{g}}\vert\text{BoN}_{\theta}(\bm{g})-\text{BoN}_{\hat{\bm{Y}}}(\bm{g})\vert\\
&=\sum_{\bm{g}}(\text{BoN}_{\theta}(\bm{g})+\text{BoN}_{\hat{\bm{Y}}}(\bm{g})-2\min(\text{BoN}_{\theta}(\bm{g}),\text{BoN}_{\hat{\bm{Y}}}(\bm{g}))\\
&=2(T-n+1-\sum_{\bm{g}}\min(\text{BoN}_{\theta}(\bm{g}),\text{BoN}_{\hat{\bm{Y}}}(\bm{g}))).
\end{aligned}
\end{equation}

The minimum between $\text{BoN}_{\theta}(\bm{g})$ and $\text{BoN}_{\hat{\bm{Y}}}(\bm{g})$ can be understood as the number of matches for the n-gram $\bm{g}$, and the $L_1$ distance measures the number of n-grams predicted by NAT that fails to match the reference sentence. Notice that the minimum will be nonzero only if the n-gram $\bm{g}$ appears in the reference sentence. Hence we can only focus on n-grams in the reference, which significantly reduce the computational effort and storage requirement. Algorithm \ref{alg} illustrates the calculation process of $\text{BoN-}L_1$.

We normalize the $L_1$ distance to range $[0,1]$ by dividing the constant $2(T-n+1)$. The BoN training objective is to minimize the normalized $L_1$ distance:
\begin{equation}
\label{eq:loss}
\mathcal{L}_{BoN}(\theta) = \frac{\text{BoN-}L_1}{2(T-n+1)}.
\end{equation}

\begin{algorithm}[t]
\caption{BoN-$L_1$} 
\label{alg}
\hspace*{0.02in} {\bf Input:} 
model parameters $\theta$, input sentence $\bm{X}$, reference sentence $\hat{\bm{Y}}$, prediction length $T$, $n$ \\
\hspace*{0.02in} {\bf Output:} 
BoN precision BoN-p
\begin{algorithmic}[1]
\State construct the reference bag-of-ngrams $\text{BoN}_{\hat{\bm{Y}}}$
\State ref-ngrams=\{$\bm{g}|\text{BoN}_{\hat{\bm{Y}}}(\bm{g})$ != $0$\}
\State match = $0$
\For{$\bm{g}$ in ref-ngrams}
  \State calculate $\text{BoN}_{\theta}(\bm{g})$ according to Eq.(\ref{eq:bon_theta_reduce})
  \State match +=  $\min(\text{BoN}_{\theta}(\bm{g}),\text{BoN}_{\hat{\bm{Y}}}(\bm{g}))$
\EndFor
\State BoN-$L_1$ = 2(T-n+1-match)
\State \Return BoN-$L_1$
\end{algorithmic}
\end{algorithm}

As is usual practice in sequence-level training, we can use the BoN objective to fine-tune a pre-trained baseline model. We denote this method as \textbf{BoN-FT}. In addition, since the BoN objective is differentiable and can be calculated efficiently, it can also work together with cross-entropy loss to jointly train NAT from scratch. We use a hyper-parameter $\alpha$ to combine the two losses:
\begin{equation}
\label{eq:joint_loss}
\mathcal{L}_{joint}(\theta) = \alpha \cdot \mathcal{L}_{MLE}(\theta) + (1-\alpha) \cdot \mathcal{L}_{BoN}(\theta).
\end{equation}
The joint training method is denoted as \textbf{BoN-Joint}. After the joint training, we can fine-tune the model using the BoN objective only. We denote this method as \textbf{BoN-Joint+FT}.
\section{Related Work}
\citet{gu2017non} introduced the non-autoregressive Transformer to reduce the translation latency of NMT, which comes at the cost of translation quality. Instead of using previously generated tokens as in autoregressive models, NAT take other global signals as decoder inputs. ~\citet{gu2017non} introduced the uniform copy and fertility to copy from source inputs. ~\citet{kaiser2018fast,roy2018theory,Ma_2019} proposed to use latent variables to improve the performance of non-autoregressive Transformer. ~\citet{akoury2019syntactically} introduced syntactically supervised Transformers, which first autoregressively predicts a chunked parse tree and then generate all target tokens conditioned on it. ~\citet{lee2018deterministic} proposed to iteratively refine the translation where the outputs of decoder are fed back as inputs in the next iteration. ~\citet{Li_2019} proposed to improve non-autoregressive models through distilling knowledge from autoregressive models. ~\citet{gu2019levenshtein} introduced Levenshtein Transformer for more flexible and amenable sequence generation. \citet{wang2018semi} introduced the semi-autoregressive Transformer that generates a group of words each time. ~\citet{guo2019non} proposed to enhance decoder inputs with phrase-table lookup and embedding mapping. ~\citet{libovicky2018end} proposed an end-to-end non-autoregressive model using connectionist temporal classification. ~\citet{wei-etal-2019-imitation} proposed an imitation learning framework where the non-autoregressive learner learns from the autoregressive demonstrator through an imitation module. ~\citet{wang2019non} pointed out that NAT is weak in capturing sequential dependency and proposed the similarity regularization and reconstruction regularization to reduce errors of repeated and incomplete translations. ~\citet{anonymous2020pnat,ran2019guiding} proposed to model the reordering information for non-autoregressive Transformer. ~\citet{shao2019retrieving} proposed Reinforce-NAT to train NAT with sequence-level objectives. As the techniques for variance reduction makes the training speed relatively slow, this method is restricted in fine-tuning. 

Training NMT with discrete sequence-level objectives has been widely investigated under the reinforcement learning framework~\cite{ranzato2015sequence,shen2016minimum,wu2016google,wu2018study,shao2019retrieving}. Recently, differentiable sequence-level training objectives for autoregressive NMT have attracted much attention. ~\citet{gu2017trainable} applied the deterministic policy gradient algorithm to train NMT actor with the differentiable critic. ~\citet{ma2018bag} proposed to use bag-of-words as target during training. ~\citet{shao2018greedy} proposed to evaluate NMT outputs with probabilistic n-grams. 
\section{Experimental Setting}
\subsubsection{Datasets}
We use several widely adopted benchmark datasets to evaluate the effectiveness of our proposed method: IWSLT16 En$\rightarrow$De (196k pairs), WMT14 En$\leftrightarrow$De (4.5M pairs) and WMT16 En$\leftrightarrow$Ro (610k pairs). For WMT14 En$\leftrightarrow$De, we employ \texttt{newstest-2013} and \texttt{newstest-2014} as development and test sets. For WMT16 En$\leftrightarrow$Ro, we take \texttt{newsdev-2016} and \texttt{newstest-2016} as development and test sets. For IWSLT16 En$\rightarrow$De, we use the \texttt{test2013} for validation. We use the preprocessed datasets released by~\citet{lee2018deterministic}, where all sentences are tokenized and segmented into subwords units~\cite{sennrich2015neural}. The vocabulary size is 40k and is shared for source and target languages. We use BLEU~\cite{papineni2002bleu} to evaluate the translation quality. 

\subsubsection{Baselines}
We take the Transformer model~\cite{vaswani2017attention} as our autoregressive baseline as well as the teacher model. The non-autoregressive model with 2 iterative refinement iterations (IRNAT)~\cite{lee2018deterministic} is our non-autoregressive baseline, and the methods we propose are all experimented on this baseline model. We also include three relevant NAT works for comparison, NAT with fertility (NAT-FT)~\cite{gu2017non}, NAT with auxiliary regularization (NAT-REG)~\cite{wang2019non} and reinforcement learning for NAT (Reinforce-NAT)~\cite{shao2019retrieving}. 

\subsubsection{Model Configurations}
We closely follow the settings of ~\citet{gu2017non}, ~\citet{lee2018deterministic} and ~\citet{shao2019retrieving}. For IWSLT16 En$\rightarrow$De, we use the small Transformer ($d_{\rm model}$=$278$, $d_{\rm hidden}$=$507$, $n_{\rm layer}$=$5$, $n_{\rm head}$=$2$, $p_{\rm dropout}$=$0.1$, $t_{\rm warmup}$=$746$). For experiments on WMT datasets, we use the base Transformer~\cite{vaswani2017attention} ($d_{\rm model}$=$512$, $d_{\rm hidden}$=$512$, $n_{\rm layer}$=$6$, $n_{\rm head}$=$8$, $p_{\rm dropout}$=$0.1$, $\rm t_{warmup}$=$16000$). We use Adam~\cite{DBLP:journals/corr/KingmaB14} for the optimization. In the main experiment, the hyper-parameter $\alpha$ to combine the BoN objective and cross-entropy loss set to be $0.1$. We set $n$=2, that is, we use the bag-of-2grams objective to train the model. The effects of $\alpha$ and $n$ will be analyzed in detail later.
{
\centering
\begin{table*}[t]
  \begin{center}
    \begin{tabular}[b]{llrrr|rrr|rrr}
    \toprule
    & & \multicolumn{3}{c|}{IWSLT'16 En-De} & \multicolumn{3}{c|}{WMT'16 En-Ro} & \multicolumn{3}{c}{WMT'14 En-De}  \\
    & & En$\rightarrow$  & speedup & secs/b & En$\rightarrow$ & Ro$\rightarrow$  & speedup  & En$\rightarrow$ & De$\rightarrow$  & speedup  \\
    \midrule
    \multirow{2}{*}{\rotatebox[origin=c]{90}{\scriptsize AR}}
        & b=1   & 28.64    &  1.09$\times$ &  0.20 & 31.93 & 31.55  &  1.23$\times$  & 23.77  & 28.15  &  1.13$\times$ \\
     & b=4   & 28.98   &  1.00$\times$ & 0.20 & 32.40 & 32.06  &  1.00$\times$ & 24.57  & 28.47 &  1.00$\times$ \\  
    \midrule
    \multirow{5}{*}{\rotatebox[origin=c]{90}{\scriptsize NAT Models}}
        & NAT-FT~\cite{gu2017non} & 26.52  & 15.6$\times$ & -- & 27.29 & 29.06 & -- & 17.69 & 21.47 & -- \\
        & IRNAT(iter=2)~\shortcite{lee2018deterministic}       & 24.82  &  6.64$\times$ & -- & 27.10 & 28.15  &  7.68$\times$  & 16.95 & 20.39 &   8.77$\times$  \\

        & IRNAT(adaptive)~\shortcite{lee2018deterministic}       & 27.01  &  1.97$\times$ & -- & 29.66 & 30.30  &  2.73$\times$  & 21.54 & 25.43 &   2.38$\times$  \\

        &NAT-REG~\cite{wang2019non}& --   &  -- & --  & --  &--   &  --  &20.65&24.77& 27.6$\times$\\
        & Reinforce-NAT~\cite{shao2019retrieving}   & 25.18  &  8.43$\times$ &13.40 & 27.09  & 27.93 &  9.44$\times$ & 19.15 & 22.52 &  10.73$\times$\\

    \midrule
     \multirow{4}{*}{\rotatebox[origin=c]{90}{\scriptsize Our Models}}
     & NAT-Base   & 24.13   &  8.42$\times$& 0.62 & 25.96  & 26.49 &  9.41$\times$  & 16.05 & 19.46  &  10.76$\times$ \\
     & BoN-FT ($n$=2) & 25.03   &  8.44$\times$& 1.41 & 27.21  & 27.95 &  9.50$\times$  & 19.27 & 23.20  &  10.72$\times$ \\
     & BoN-Joint ($n$=2, $\alpha$=0.1) & 25.63   &  8.39$\times$& 1.49 & 28.12   & 29.03 &  9.44$\times$   & 20.75  & 24.47 &  10.79$\times$\\
     & BoN-Joint+FT ($n$=2, $\alpha$=0.1)& 25.72   &  8.40$\times$& 1.41 & 28.31   & 29.29 &  9.51$\times$  & 20.90  & 24.61   &  10.77$\times$ \\
    \bottomrule
  \end{tabular}
  
  \caption{Generation quality (4-gram BLEU), speedup and training speed (seconds/batch). Decoding speed is measured sentence-by-sentence from the En$\rightarrow$ direction. AR: the autoregressive Transformer baseline and the teacher model. $b$: beam size. NAT-Base: the non-autoregressive Transformer baseline trained by the cross-entropy loss. BoN-FT: fine-tune the NAT-Base with the BoN objective. BoN-Joint: combine the BoN objective and cross-entropy loss to jointly train NAT from scratch. BoN-Joint+FT: fine-tune the BoN-Joint with the BoN objective.}
  \label{tab:bleu_performance}
  \end{center}
\end{table*}
}
\subsubsection{Training and Inference}
During training, we apply the sequence-level knowledge distillation~\cite{kim2016sequence}, which constructs the distillation corpus where the target side of the training corpus is replaced by the output of the autoregressive Transformer. We directly use the distillation corpus released by ~\citet{lee2018deterministic}. We use the original corpus to train the autoregressive Transformer and distillation corpus to train non-autoregressive models. For NAT baseline, we stop training when the number of training steps exceeds 300k and there is no further improvements in the last 100k steps. For BoN-Joint, we stop training when the number of training steps exceeds 150k and there is no further improvements in the last 100k steps. For BoN-FT, the number of fine-tuning steps is fixed to 5k.

After training NAT, we train a target length predictor to predict the length difference between the source and target sentences. The target length predictor takes the sum of encoder hidden states as input and feeds it to a softmax classifier after an affine transformation. 
During inference, we first run the encoder and apply the length predictor to predict the target length. Then we construct decoder inputs through uniform copy~\cite{gu2017non} and run the decoder to generate the translation. During postprocessing, we remove any token that is generated repeatedly. The training and decoding speed are measured on a single Geforce GTX TITAN X.

\section{Results and Analysis}
\subsection{Main Results}
We report our main results in Table \ref{tab:bleu_performance}, from which we have the following observations:

\noindent
1. {\bf BoN-FT achieves significant training speedup over Reinforce-NAT and outperforms Reinforce-NAT on BLEU scores}. BoN-FT and Reinforce-NAT are both fine-tuning approaches. BoN-FT slightly outperforms Reinforce-NAT on four translation directions and achieves at most 3.74 BLEU improvements over the NAT-Base on WMT14 De$\rightarrow$En. More importantly, since the BoN objective is differentiable and can be efficiently calculated, the training speed of BoN-FT is nearly 10 times faster than Reinforce-NAT, which makes the fine-tuning process very efficient. 

\noindent
2. {\bf BoN-Joint+FT achieves considerable improvements over BoN-FT}. By combining the BoN objective and cross-entropy loss, we obtain BoN-Joint, which achieves considerable improvements over BoN-FT. Though the training speed for BoN-Joint is about two times slower, BoN-Joint only needs about half of training steps to converge, so the training time is almost the same as NAT-Base. BoN-Joint+FT further improves the performance through fine-tuning BoN-Joint with the BoN objective, which achieves about 5.0 BLEU improvements on WMT14 En$\leftrightarrow$De, about 2.5 BLEU improvements on WMT16 En$\leftrightarrow$Ro and 1.59 BLEU improvements on IWSLT16 En$\rightarrow$De. 

\noindent
3. {\bf Our methods achieve higher improvements on WMT datasets than the IWSLT dataset}. One interesting phenomenon is that our methods achieve higher improvements on WMT datasets. The BLEU improvement is 4.85 on WMT14 En$\rightarrow$De but only 1.59 on IWSLT16 En$\rightarrow$De. As we will analyze later, this has a lot to do with the sentence length of datasets since the cross-entropy loss does not perform well on long sentences.

\subsection{Correlation with Translation Quality}

In this section, we experiment with the correlation between the loss function and translation quality. We are interested in how the cross-entropy loss and BoN objective correlate with the translation quality. As a common practice, the BLEU score is used to represent the translation quality. We experiment on the WMT14 En$\rightarrow$De development set, which contains 3000 sentences. Firstly we randomly divide the dataset into 100 subsets of size 30. Then we use the NAT model to decode the 100 subsets and compute the BLEU score and loss function on every subset. Finally we calculate the pearson correlation between the 100 BLEU scores and losses.

For the cross-entropy loss, we normalize it by the target sentence length. The BoN training objective is the $L_1$ distance BoN-$L_1$ normalized by $2(T-n+1)$. We respectively set $n$ to $1$, $2$, $3$ and $4$ to test different n-gram size. Table \ref{tab:n} lists the correlation results.
\begin{table}[!htbp]
\centering
\resizebox{.95\columnwidth}{!}{
\begin{tabular}{c|c|c|c|c|c}
\toprule
Loss function& CE& $n=1$ & $n=2$ &$n=3$ &$n=4$\\
\hline
Correlation&0.37&0.56&0.70&0.67&0.61\\
\bottomrule
\end{tabular}}
\caption{The pearson correlation bewteen loss functions and translation quality. $n=k$ represents the bag-of-kgrams training objective. CE represents the cross-entropy loss.}
\label{tab:n}
\end{table}

We can see that between different BoN objectives, $n=1$, which represents the bag-of-words objective, underperforms other choices of $n$, indicating the necessity of modeling sequential dependency for NAT. Another observation is that all the four BoN objectives outperform the cross-entropy loss by large margins. To find out where the performance gap comes from, we analyze the effect of sentence length in the following experiment. We evenly divide the dataset into two parts according to the source length, where the first part consists of 1500 short sentences and the second part consists of 1500 long sentences. We respectively measure the pearson correlation on the two parts and list the results in Table \ref{tab:joint}:
\begin{table}[!htbp]
\centering
\begin{tabular}{c|c|c|c}
\toprule
& all& short& long\\
\hline
Cross-Entropy&0.37&0.52&0.21\\
\hline
BoN ($n$=2)&0.70&0.79&0.81\\
\bottomrule
\end{tabular}
\caption{The pearson correlation bewteen loss functions and translation quality on short sentences and long sentences.}
\label{tab:joint}
\end{table}

From Table \ref{tab:joint}, we can see that the correlation of cross-entropy loss drops sharply as the sentence length increases, where the BoN objective still has strong correlation on long sentences. The reason is not difficult to explain. The cross-entropy loss requires the strict alignment between the translation and reference. As sentence length grows, it becomes harder for NAT to align the translation with reference, which leads to a decrease of correlation between cross-entropy loss and translation quality. In contrast, the BoN objective is robust to unaligned translations, so its correlation with translation quality stays strong when translating long sentences.

\subsection{Number of Removed Tokens}
In this section, we study the number of tokens removed during postprocessing to see how our method improves translation quality. We sort the WMT14 En$\rightarrow$De development set by sentence length and split it evenly into the short part and long part accordingly. Then we translate them using NAT-Base and BoN-Joint respectively and record in table \ref{tab:remove} the number of removed repeated tokens during postprocessing.
\begin{table}[!htbp]
\centering
\resizebox{.95\columnwidth}{!}{
\begin{tabular}{c|c|c|c}
\toprule
& short& long& all\\
\hline
Total&22023&56142&78165\\
\hline
NAT-Base&1232(5.6\%)&7317(13.0\%)&9643(12.3\%)\\
\hline
BoN-Joint&515(2.3\%)&1726(3.1\%)&2241(2.9\%)\\
\bottomrule
\end{tabular}}
\caption{The number of removed repeated tokens for NAT-Base and BoN-Joint. Total represents the total length of reference sentences.}
\label{tab:remove}
\end{table}

As shown in table \ref{tab:remove}, NAT-Base suffers from the repeated translation when translating long sentences, which is efficiently alleviated in BoN-Joint. This is in line with our analysis. As sentence length grows, it becomes harder for NAT to align the translation with reference, which results in more overcorrection errors and makes more repeated tokens be translated.

\subsection{Effect of Sentence Length}
\begin{figure}[ht]
  \begin{center}
    \includegraphics[width=.95\columnwidth]{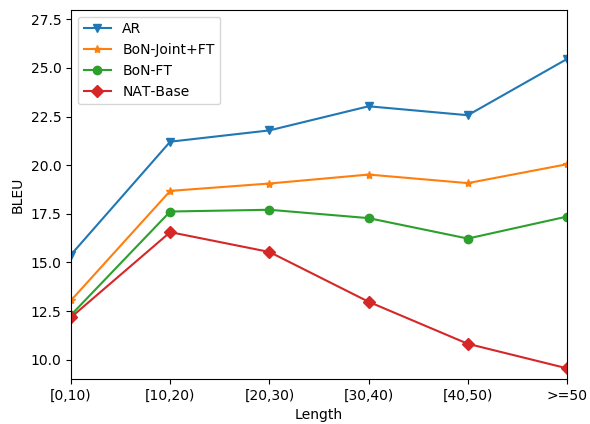}
    \caption{BLEU performance on different length buckets.}
    \label{fig:length}
  \end{center}
\end{figure}

Previously, we give the correlation between loss functions and the translation quality under different sentence lengths. In this section, we will experiment on the BLEU performance of NAT baseline and our models on different sentence lengths and see whether the better correlation contributes to better BLEU performance. We conduct the analysis on the WMT14 En$\rightarrow$De development set and divide the sentence pairs into different length buckets according to the length of the reference sentence. Figure \ref{fig:length} shows our results. 

We can see that the performance of NAT-Base drops quickly as sentence length increases, where the autoregressive Transformer and BoN models have stable performance over different sentence lengths. This can be well explained by the correlation results. As sentence length grows, the correlation between the cross-entropy loss and the translation quality drops sharply, which leads to the weakness of NAT in translating long sentences.

\subsection{Effect of $\alpha$}
\begin{figure}[ht]
  \begin{center}
    \includegraphics[width=.95\columnwidth]{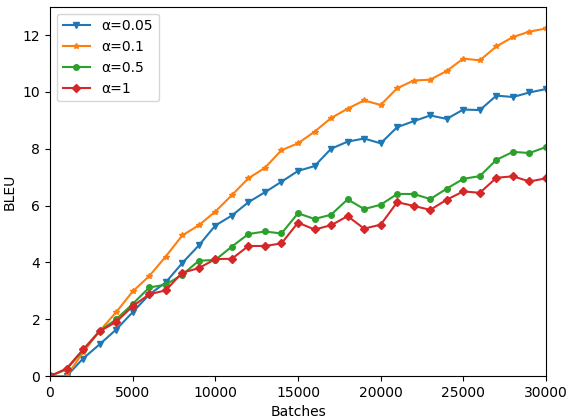}
    \caption{Training curves for BoN-Joint with different hyper-parameter $\alpha$.}
    \label{fig:alpha}
  \end{center}
\end{figure}

In this section, we study the effect of the hyper-parameter $\alpha$, which is defined in Eq.(\ref{eq:joint_loss}) and is used to control the proportion of the cross-entropy loss and BoN loss. We set $n=2$ and use different $\alpha$ to train BoN-Joint on WMT14 En$\rightarrow$De. Figure \ref{fig:alpha} shows the BLEU scores on the development set.

We can see from Figure \ref{fig:alpha} that $\alpha=0.1$ has a stable and fast BLEU improvement over time. When we set $\alpha=0.1$, the time required for model convergence is the shortest, and the BLEU score after convergence is the best over other choices of $\alpha$. In addition, we find that when we set $n=2$, $\alpha=0.1$ performs well in all the three datasets, which frees us from the hyper-parameter tuning. 
\subsection{Effect of N-gram Size}

In this section, we study the effect of n-gram size. We respectively set $n$ to $1$, $2$, $3$, and $4$ and evaluate the performance of BoN-FT and BoN-Joint on the WMT14 En$\rightarrow$De development set. When training BoN-Joint, we tune the hyper-parameter $\alpha$ for every $n$ to obtain the optimal performance. The results are listed in Table \ref{tab:bleu_n}. For BoN-FT, the BoN objective is only used to fine-tune the NAT baseline, and different choices for $n$ do not have a large impact on the BLEU performance. For BoN-Joint, the choice of $n$ becomes more important, and $n=2$ significantly outperforms other choices of $n$, which is consistent with the correlation result in Table \ref{tab:n}.

\begin{table}[!ht]
\centering
\begin{tabular}{c|c|c|c|c}
\toprule
&  $n=1$ & $n=2$ &$n=3$ &$n=4$\\
\hline
BoN-FT&18.06&18.13&18.17&18.08\\
\hline
BoN-Joint&17.73&19.28&18.45&17.97\\
\bottomrule
\end{tabular}
\caption{The BLEU performance of BoN-FT and BoN-Joint under different choices of $n$.}
\label{tab:bleu_n}
\end{table}

\section{Conclusion}
In the context of non-autoregressive translation, the cross-entropy loss cannot model the target-side sequential dependency properly and suffers from the weak correlation with the translation quality. In this paper, we propose a novel Bag-of-Ngrams objective to model the target-side sequential dependency for NAT. The BoN objective is differentiable and can be calculated efficiently. Experiments show that the BoN objective correlates well with the translation quality and achieves large improvements over the NAT baseline. 
\section{Acknowledgments}
We thank the anonymous reviewers for their insightful comments. This work was supported by National Natural Science Foundation of China (NO.61876174) and National Key R\&D Program of China (NO.2017YFE9132900).

\bibliography{AAAI-ShaoC.1194}

\begin{thebibliography}{}

\bibitem[\protect\citeauthoryear{Akoury, Krishna, and
  Iyyer}{2019}]{akoury2019syntactically}
Akoury, N.; Krishna, K.; and Iyyer, M.
\newblock 2019.
\newblock Syntactically supervised transformers for faster neural machine
  translation.
\newblock In {\em ACL}.

\bibitem[\protect\citeauthoryear{Anonymous}{2020}]{anonymous2020pnat}
Anonymous.
\newblock 2020.
\newblock {\{}PNAT{\}}: Non-autoregressive transformer by position learning.
\newblock In {\em Submitted to International Conference on Learning
  Representations}.
\newblock under review.

\bibitem[\protect\citeauthoryear{Bahdanau, Cho, and
  Bengio}{2014}]{bahdanau2014neural}
Bahdanau, D.; Cho, K.; and Bengio, Y.
\newblock 2014.
\newblock Neural machine translation by jointly learning to align and
  translate.
\newblock {\em arXiv preprint arXiv:1409.0473}.

\bibitem[\protect\citeauthoryear{Cho \bgroup et al\mbox.\egroup
  }{2014}]{cho2014learning}
Cho, K.; Van~Merri{\"e}nboer, B.; Gulcehre, C.; Bahdanau, D.; Bougares, F.;
  Schwenk, H.; and Bengio, Y.
\newblock 2014.
\newblock Learning phrase representations using rnn encoder-decoder for
  statistical machine translation.
\newblock {\em arXiv preprint arXiv:1406.1078}.

\bibitem[\protect\citeauthoryear{Gehring \bgroup et al\mbox.\egroup
  }{2017}]{gehring2017convolutional}
Gehring, J.; Auli, M.; Grangier, D.; Yarats, D.; and Dauphin, Y.~N.
\newblock 2017.
\newblock Convolutional sequence to sequence learning.
\newblock {\em arXiv preprint arXiv:1705.03122}.

\bibitem[\protect\citeauthoryear{Gu \bgroup et al\mbox.\egroup
  }{2017}]{gu2017non}
Gu, J.; Bradbury, J.; Xiong, C.; Li, V.~O.; and Socher, R.
\newblock 2017.
\newblock Non-autoregressive neural machine translation.
\newblock In {\em ICLR}.

\bibitem[\protect\citeauthoryear{Gu, Cho, and Li}{2017}]{gu2017trainable}
Gu, J.; Cho, K.; and Li, V.~O.
\newblock 2017.
\newblock Trainable greedy decoding for neural machine translation.
\newblock In {\em EMNLP}.

\bibitem[\protect\citeauthoryear{Gu, Wang, and Zhao}{2019}]{gu2019levenshtein}
Gu, J.; Wang, C.; and Zhao, J.
\newblock 2019.
\newblock Levenshtein transformer.
\newblock In {\em Advances in NeurIPS}.

\bibitem[\protect\citeauthoryear{Guo \bgroup et al\mbox.\egroup
  }{2019}]{guo2019non}
Guo, J.; Tan, X.; He, D.; Qin, T.; Xu, L.; and Liu, T.-Y.
\newblock 2019.
\newblock Non-autoregressive neural machine translation with enhanced decoder
  input.
\newblock In {\em AAAI}, volume~33.

\bibitem[\protect\citeauthoryear{Joachims}{1998}]{joachims1998text}
Joachims, T.
\newblock 1998.
\newblock Text categorization with support vector machines: Learning with many
  relevant features.
\newblock In {\em European conference on machine learning}.
\newblock Springer.

\bibitem[\protect\citeauthoryear{Kaiser \bgroup et al\mbox.\egroup
  }{2018}]{kaiser2018fast}
Kaiser, L.; Bengio, S.; Roy, A.; Vaswani, A.; Parmar, N.; Uszkoreit, J.; and
  Shazeer, N.
\newblock 2018.
\newblock Fast decoding in sequence models using discrete latent variables.
\newblock In {\em ICML}.

\bibitem[\protect\citeauthoryear{Kim and Rush}{2016}]{kim2016sequence}
Kim, Y., and Rush, A.~M.
\newblock 2016.
\newblock Sequence-level knowledge distillation.
\newblock In {\em EMNLP}.

\bibitem[\protect\citeauthoryear{Kingma and
  Ba}{2014}]{DBLP:journals/corr/KingmaB14}
Kingma, D.~P., and Ba, J.
\newblock 2014.
\newblock Adam: {A} method for stochastic optimization.
\newblock {\em CoRR} abs/1412.6980.

\bibitem[\protect\citeauthoryear{Lee, Mansimov, and
  Cho}{2018}]{lee2018deterministic}
Lee, J.; Mansimov, E.; and Cho, K.
\newblock 2018.
\newblock Deterministic non-autoregressive neural sequence modeling by
  iterative refinement.
\newblock In {\em EMNLP}.

\bibitem[\protect\citeauthoryear{Li \bgroup et al\mbox.\egroup
  }{2017}]{li2017neural}
Li, B.; Liu, T.; Zhao, Z.; Wang, P.; and Du, X.
\newblock 2017.
\newblock Neural bag-of-ngrams.
\newblock In {\em AAAI}.

\bibitem[\protect\citeauthoryear{Li \bgroup et al\mbox.\egroup
  }{2019}]{Li_2019}
Li, Z.; Lin, Z.; He, D.; Tian, F.; QIN, T.; WANG, L.; and Liu, T.-Y.
\newblock 2019.
\newblock Hint-based training for non-autoregressive machine translation.
\newblock In {\em EMNLP-IJCNLP}.

\bibitem[\protect\citeauthoryear{Libovick{\`y} and
  Helcl}{2018}]{libovicky2018end}
Libovick{\`y}, J., and Helcl, J.
\newblock 2018.
\newblock End-to-end non-autoregressive neural machine translation with
  connectionist temporal classification.
\newblock In {\em EMNLP}.

\bibitem[\protect\citeauthoryear{Ma \bgroup et al\mbox.\egroup
  }{2018}]{ma2018bag}
Ma, S.; Xu, S.; Wang, Y.; and Lin, J.
\newblock 2018.
\newblock Bag-of-words as target for neural machine translation.
\newblock In {\em ACL}.

\bibitem[\protect\citeauthoryear{Ma \bgroup et al\mbox.\egroup
  }{2019}]{Ma_2019}
Ma, X.; Zhou, C.; Li, X.; Neubig, G.; and Hovy, E.
\newblock 2019.
\newblock Flowseq: Non-autoregressive conditional sequence generation with
  generative flow.
\newblock In {\em EMNLP-IJCNLP}.

\bibitem[\protect\citeauthoryear{Pang, Lee, and
  Vaithyanathan}{2002}]{pang2002thumbs}
Pang, B.; Lee, L.; and Vaithyanathan, S.
\newblock 2002.
\newblock Thumbs up?: sentiment classification using machine learning
  techniques.
\newblock In {\em ACL}.

\bibitem[\protect\citeauthoryear{Papineni \bgroup et al\mbox.\egroup
  }{2002}]{papineni2002bleu}
Papineni, K.; Roukos, S.; Ward, T.; and Zhu, W.-J.
\newblock 2002.
\newblock Bleu: a method for automatic evaluation of machine translation.
\newblock In {\em ACL}.

\bibitem[\protect\citeauthoryear{Ran \bgroup et al\mbox.\egroup
  }{2019}]{ran2019guiding}
Ran, Q.; Lin, Y.; Li, P.; and Zhou, J.
\newblock 2019.
\newblock Guiding non-autoregressive neural machine translation decoding with
  reordering information.
\newblock {\em arXiv preprint arXiv:1911.02215}.

\bibitem[\protect\citeauthoryear{Ranzato \bgroup et al\mbox.\egroup
  }{2015}]{ranzato2015sequence}
Ranzato, M.; Chopra, S.; Auli, M.; and Zaremba, W.
\newblock 2015.
\newblock Sequence level training with recurrent neural networks.
\newblock {\em arXiv preprint arXiv:1511.06732}.

\bibitem[\protect\citeauthoryear{Roy \bgroup et al\mbox.\egroup
  }{2018}]{roy2018theory}
Roy, A.; Vaswani, A.; Neelakantan, A.; and Parmar, N.
\newblock 2018.
\newblock Theory and experiments on vector quantized autoencoders.
\newblock {\em arXiv preprint arXiv:1805.11063}.

\bibitem[\protect\citeauthoryear{Sennrich, Haddow, and
  Birch}{2016}]{sennrich2015neural}
Sennrich, R.; Haddow, B.; and Birch, A.
\newblock 2016.
\newblock Neural machine translation of rare words with subword units.
\newblock In {\em ACL}.

\bibitem[\protect\citeauthoryear{Shao \bgroup et al\mbox.\egroup
  }{2019}]{shao2019retrieving}
Shao, C.; Feng, Y.; Zhang, J.; Meng, F.; Chen, X.; and Zhou, J.
\newblock 2019.
\newblock Retrieving sequential information for non-autoregressive neural
  machine translation.
\newblock In {\em ACL}.

\bibitem[\protect\citeauthoryear{Shao, Chen, and Feng}{2018}]{shao2018greedy}
Shao, C.; Chen, X.; and Feng, Y.
\newblock 2018.
\newblock Greedy search with probabilistic n-gram matching for neural machine
  translation.
\newblock In {\em EMNLP}.

\bibitem[\protect\citeauthoryear{Shen \bgroup et al\mbox.\egroup
  }{2015}]{shen2016minimum}
Shen, S.; Cheng, Y.; He, Z.; He, W.; Wu, H.; Sun, M.; and Liu, Y.
\newblock 2015.
\newblock Minimum risk training for neural machine translation.
\newblock In {\em ACL}.

\bibitem[\protect\citeauthoryear{Sutskever, Vinyals, and
  Le}{2014}]{sutskever2014sequence}
Sutskever, I.; Vinyals, O.; and Le, Q.~V.
\newblock 2014.
\newblock Sequence to sequence learning with neural networks.
\newblock In {\em Advances in NIPS}.

\bibitem[\protect\citeauthoryear{Vaswani \bgroup et al\mbox.\egroup
  }{2017}]{vaswani2017attention}
Vaswani, A.; Shazeer, N.; Parmar, N.; Uszkoreit, J.; Jones, L.; Gomez, A.~N.;
  Kaiser, {\L}.; and Polosukhin, I.
\newblock 2017.
\newblock Attention is all you need.
\newblock In {\em Advances in NIPS}.

\bibitem[\protect\citeauthoryear{Wang \bgroup et al\mbox.\egroup
  }{2019}]{wang2019non}
Wang, Y.; Tian, F.; He, D.; Qin, T.; Zhai, C.; and Liu, T.-Y.
\newblock 2019.
\newblock Non-autoregressive machine translation with auxiliary regularization.
\newblock In {\em AAAI}.

\bibitem[\protect\citeauthoryear{Wang, Zhang, and Chen}{2018}]{wang2018semi}
Wang, C.; Zhang, J.; and Chen, H.
\newblock 2018.
\newblock Semi-autoregressive neural machine translation.
\newblock In {\em EMNLP}.

\bibitem[\protect\citeauthoryear{Wei \bgroup et al\mbox.\egroup
  }{2019}]{wei-etal-2019-imitation}
Wei, B.; Wang, M.; Zhou, H.; Lin, J.; and Sun, X.
\newblock 2019.
\newblock Imitation learning for non-autoregressive neural machine translation.
\newblock In {\em ACL}.

\bibitem[\protect\citeauthoryear{Williams}{1992}]{williams1992simple}
Williams, R.~J.
\newblock 1992.
\newblock Simple statistical gradient-following algorithms for connectionist
  reinforcement learning.
\newblock In {\em Reinforcement Learning}. Springer.

\bibitem[\protect\citeauthoryear{Wu \bgroup et al\mbox.\egroup
  }{2016}]{wu2016google}
Wu, Y.; Schuster, M.; Chen, Z.; Le, Q.~V.; Norouzi, M.; Macherey, W.; Krikun,
  M.; Cao, Y.; Gao, Q.; Macherey, K.; et~al.
\newblock 2016.
\newblock Google's neural machine translation system: Bridging the gap between
  human and machine translation.
\newblock {\em arXiv preprint arXiv:1609.08144}.

\bibitem[\protect\citeauthoryear{Wu \bgroup et al\mbox.\egroup
  }{2018}]{wu2018study}
Wu, L.; Tian, F.; Qin, T.; Lai, J.; and Liu, T.-Y.
\newblock 2018.
\newblock A study of reinforcement learning for neural machine translation.
\newblock In {\em EMNLP}.

\bibitem[\protect\citeauthoryear{Zhang \bgroup et al\mbox.\egroup
  }{2019}]{zhang-etal-2019-bridging}
Zhang, W.; Feng, Y.; Meng, F.; You, D.; and Liu, Q.
\newblock 2019.
\newblock Bridging the gap between training and inference for neural machine
  translation.
\newblock In {\em ACL}.

\end{thebibliography}
\bibliographystyle{aaai}
\end{document}